\documentclass[conference]{IEEEtran}
\usepackage{times}
\usepackage{amsfonts}
\usepackage{subcaption}
\usepackage{graphicx}
\usepackage{url}
\usepackage{color}

\usepackage[numbers]{natbib}
\usepackage{multicol}
\usepackage[bookmarks=true]{hyperref}

\IEEEoverridecommandlockouts
\begin{document}

\definecolor{RoyalBlue}{RGB}{65,105,225}
\newcommand{\todo}[1]{\textbf{\textcolor{red}{TODO: #1}}}

\title{Reinforcement and Imitation Learning\\for Diverse Visuomotor Skills}


\author{Yuke Zhu$^\dagger$\thanks{*\,This work was done when Yuke Zhu ({\tt yukez@cs.stanford.edu}) worked as a summer intern at DeepMind.}
\qquad Ziyu Wang$^\ddagger$
\qquad Josh Merel$^\ddagger$
\qquad Andrei Rusu$^\ddagger$
\qquad Tom Erez$^\ddagger$
\qquad Serkan Cabi$^\ddagger$\\
Saran Tunyasuvunakool$^\ddagger$
\qquad J\'{a}nos Kram\'{a}r$^\ddagger$
\qquad Raia Hadsell$^\ddagger$
\qquad Nando de Freitas$^\ddagger$
\qquad Nicolas Heess$^\ddagger$\\
$^\dagger$Computer Science Department, Stanford University, USA\\
$^\ddagger$DeepMind, London, UK
}



%

\maketitle

\begin{abstract}
We propose a model-free deep reinforcement learning method that leverages a small amount of demonstration data to assist a reinforcement learning agent. We apply this approach to robotic manipulation tasks and train end-to-end visuomotor policies that map directly from RGB camera inputs to joint velocities. 
%
We demonstrate that our approach can solve a wide variety of visuomotor tasks, for which engineering a scripted controller would be laborious.
In experiments, our reinforcement and imitation agent achieves significantly better performances than agents trained with reinforcement learning or imitation learning alone.
We also illustrate that these policies, trained with large visual and dynamics variations, can achieve preliminary successes in zero-shot sim2real transfer. A brief visual description of this work can be viewed in {\color{RoyalBlue} \href{https://youtu.be/EDl8SQUNjj0}{this video}}.
\end{abstract}

\IEEEpeerreviewmaketitle

\section{Introduction}
\label{intro}
Recent advances in deep reinforcement learning (RL) have performed very well in several challenging domains such as video games~\citep{mnih2015human} and Go~\citep{silver2016mastering}. For robotics, RL in combination with powerful function approximators such as neural networks provides a general framework for designing sophisticated controllers that would be hard to handcraft otherwise. 
Reinforcement learning methods have a long history in robotics control but have typically been used with low-dimensional movement representations \citep{deisenroth2013survey,kober2012reinforcement}. The 
last few years have seen a growing number of successful demonstrations of deep RL for robotic manipulation using  model-based (e.g.\ \citet{levine2015end,yahya2016collective,levine2016learning}) and model-free techniques (e.g.\
\citet{chebotar2017path,gu2016deep,popov2017data}), 
both in simulation and on real hardware. Nevertheless, end-to-end learning of visuomotor controllers for long-horizon and multi-stage manipulation tasks using model-free RL techniques remains a challenging problem.

\begin{figure}[t]
\begin{center}
\includegraphics[width=.95\linewidth]{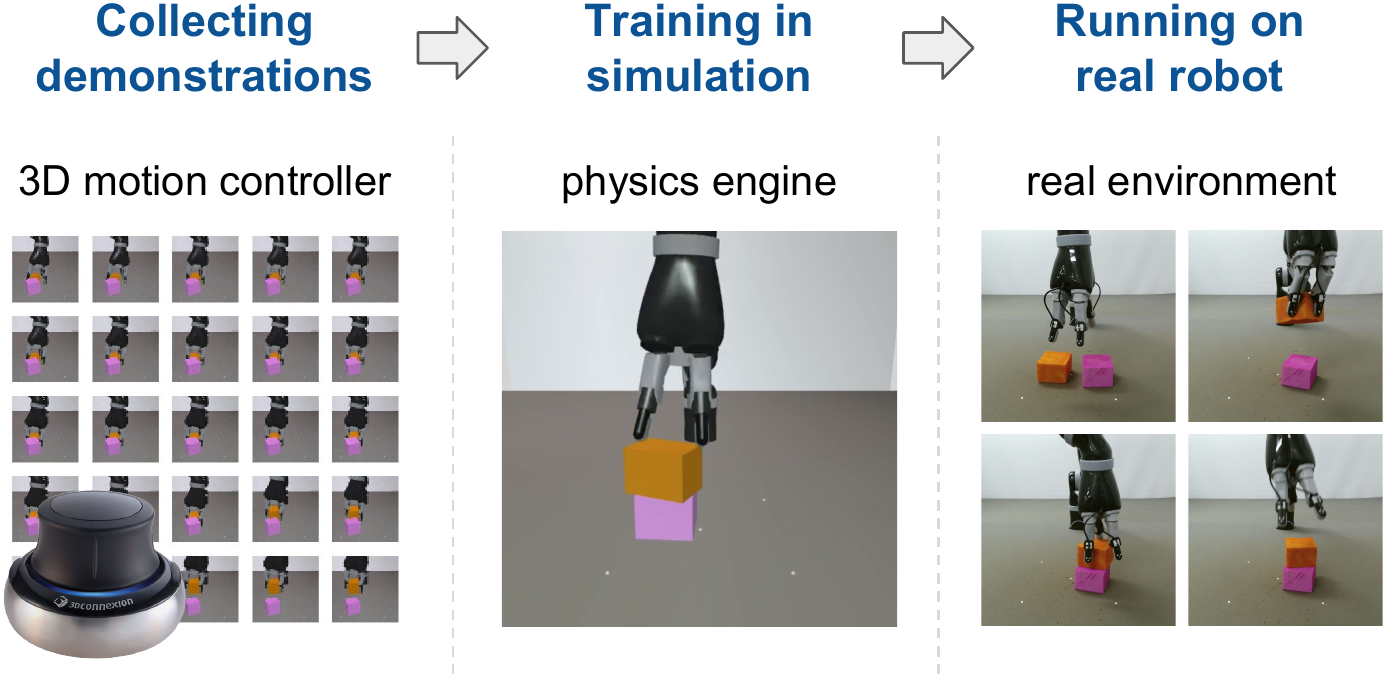}
\caption{Our proposal of a principled robot learning pipeline. We used 3D motion controllers to collect human demonstrations of a task. Our reinforcement and imitation learning model leveraged these demonstrations to facilitate learning in a simulated physical engine. We then performed sim2real transfer to deploy the learned visuomotor policy to a real robot.}
\vspace{-3mm}
\label{fig:pipeline}
\end{center}
\end{figure}


Developing RL agents for robotics requires overcoming  several significant challenges. Policies for robotics must transform multi-modal and partial observations from noisy sensors, such as cameras, into coordinated activity of many degrees of freedom.
At the same time, realistic tasks often come with contact-rich dynamics and vary along multiple dimensions (visual appearance, position, shapes, etc.), posing significant generalization challenges. Model-based methods can have difficulties handling such complex dynamics and large variations. Directly  training model-free methods on real robotics hardware can be daunting due to the high sample complexity.
The difficulty of real-world RL training is compounded by safety considerations as well as the difficulty of accessing information about the state of the environment (e.g. the position of an object) to define a reward function.
Finally, even in simulation when perfect state information and large amounts of training data are available, exploration can be a significant challenge, especially for on-policy methods. This is partly due to the often high-dimensional and continuous action space, but also due to the difficulty of designing suitable reward functions. 

In this paper, we present a model-free deep RL method 
that can solve a variety of robotic manipulation tasks directly from pixel input. Our key insights are 1) to reduce the difficulty of exploration in continuous domains by leveraging a handful of human demonstrations; 2) to leverage several new techniques 
that exploit privileged and task-specific information during training only 
which can accelerate and stabilize the learning of visuomotor policies in multi-stage tasks; and 3) to improve generalization by increasing the diversity of the training conditions. As a result, the policies work well under significant variations of system dynamics, object appearances, task lengths, etc. Furthermore, we demonstrate promising preliminary results for two tasks, where the policies trained in simulation achieve zero-shot transfer to a real robot.


We evaluate our method on six manipulation tasks, including stacking, pouring, etc. The set of tasks includes multi-stage and long-horizon tasks, and they require full 9-DoF joint velocity control directly from pixels. The controllers need to be able to handle significant shape and appearance variations.


To address these challenges, our method combines imitation learning with reinforcement learning into a unified training framework. Our approach utilizes demonstration data in two ways: first, it uses a hybrid reward that combines the task reward with an imitation reward based on Generative Adversarial Imitation Learning~\citep{ho2016generative}. This aids with exploration while still allowing the final controller to outperform the human demonstrator on the task.
Second, it uses demonstration 
trajectories to construct a curriculum of states along which
to initialize the episodes during training.
This enables the agent to learn about later stages of the task earlier in training, facilitating the solving of long tasks.
As a result, our approach solves all six tasks, which  neither the reinforcement learning nor imitation learning baselines can solve alone.

To sidestep the constraints of training on real hardware we embrace the sim2real paradigm which has recently shown promising results \citep{james2017transferring,rusu2016progressive,tobin2017domain}. Through the use of a physics engine and high-throughput RL algorithms, we can simulate parallel copies of a robot arm to perform millions of complex physical interactions in a contact-rich environment while eliminating the practical concerns of robot safety and system reset. Furthermore, we can, during training, exploit privileged and task-specific information about the true system state with several new techniques, including learning policy and value in separate modalities, an object-centric GAIL discriminator, and auxiliary tasks for visual modules. These techniques stabilize and speed up policy learning, without imposing any constraints on the system at test time. 

Finally, we diversify training conditions such as visual appearance, object geometry, and system dynamics. This improves both generalization with respect to different task conditions as well as transfer from simulation to reality.


We use the same model and the same algorithm with only small task-specific modifications of the training setup to learn visuomotor controllers for six diverse robot arm manipulation tasks.
 As illustrated in Fig.~\ref{fig:pipeline} 
 this instantiates a visuomotor learning pipeline going from collecting human demonstration to learning in simulation, and back to real-world deployment via sim2real policy transfer.

\section{Related work}
\label{related}

Reinforcement learning  methods have been extensively used with low-dimensional policy representations such as movement primitives to solve a variety of control problems both in simulation and in reality.
Three classes of RL algorithms are currently dominant for continuous control problems: guided policy search methods (GPS; \citet{levine2013guided}), value-based methods such as the deterministic policy gradient (DPG; \citet{silver2014deterministic,lillicrap2015continuous,heess2015learning}) or the normalized advantage function (NAF; \citet{gu2016continuous}) algorithm, and trust-region based policy gradient algorithms such as trust region policy optimization (TRPO) and proximal policy optimization (PPO). 
TRPO \citep{schulman2015trust} and PPO \citep{schulman2017proximal} hold appeal due to their robustness to hyperparameter settings as well as their scalability \citep{heess2017emergence} but the lack of sample efficiency makes them unsuitable for training directly on robotics hardware.

GPS \citep{levine2013guided} has been used e.g.\ by \citet{levine2015end}, \citet{yahya2016collective} and \citet{chebotar2017path} to learn visuomotor policies directly on a real robotics hardware after a network pretraining phase. \citet{gupta2016learning} 
and \citet{kumar2016learning} 
use GPS for learning controllers for robotic hand models.
Value-based methods have been employed, e.g.\ by \citet{gu2016deep} who use NAF to learn a door opening task directly on a robot while \citet{popov2017data} demonstrate how to solve a stacking problem efficiently using a distributed variant of DPG. 

The idea of using large-scale data collection for training visuomotor controllers has been the focus of \citet{levine2016learning} and \citet{pinto2015supersizing} who train a convolutional network to predict grasp success for diverse sets of objects using a large dataset with 10s or 100s of thousands of grasp attempts collected from multiple robots in a self-supervised setting.

An alternative strategy for dealing with the data demand is to train in simulation and transfer the learned controller to real hardware, or to augment real-world training with synthetic data. \citet{rusu2016simtoreal} learn simple visuomotor policies for a Jaco robot arm and transfer to reality using progressive networks~\cite{rusu2016progressive}. \citet{viereck2017learning} minimize the reality gap by relying on depth. \citet{tobin2017domain} use visual variations to learn robust object detectors that can transfer to reality; \citet{james2017transferring}  combine randomization with supervised learning. \citet{bousmalis2017using} augment the training with simulated data to learn grasp prediction of diverse shapes.

\begin{figure*}[t]
\begin{center}
\includegraphics[width=.95\linewidth]{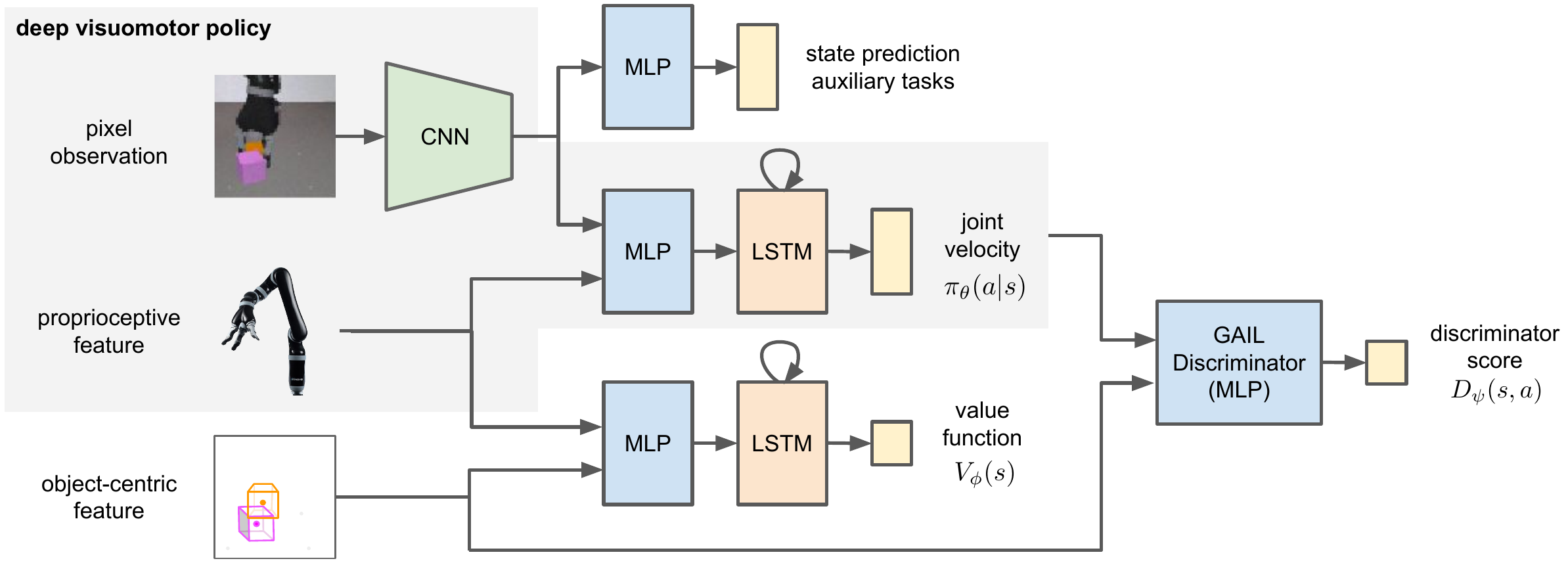}
\caption{Model overview. The core of our model is the deep visuomotor policy, which takes the camera observation and the proprioceptive feature as input and produces the next joint velocities.}
\vspace{-3mm}
\label{fig:architecture}
\end{center}
\end{figure*}

Suitable cost functions and exploration strategies for control problems are challenging to design, so demonstrations have long played an important role. Demonstrations can be used to initialize policies, design cost functions, guide exploration, augment the training data, or a combination of these. 
Cost functions can be derived from demonstrations either via tracking objectives (e.g.\ \citet{gupta2016learning}) or via inverse RL (e.g.\ \citet{boularias2011relative,finn2016guided}), or, as in our case, via adversarial learning \citep{ho2016generative}. When expert actions or expert policies are available, behavioral cloning can be used (\citet{rahmatizadeh2017vision,james2017transferring,duan2017oneshot}). Alternatively, expert trajectories can be used as additional training data for off-policy algorithms such as DPG (e.g.\ \citet{vecerik2017leveraging}). Most of these methods require observation and/or action spaces to be aligned between the robot and demonstrations. Recently, methods for third person imitation have been proposed (e.g.~\citet{sermanet2017time,liu2017imitation,finn2017oneshot}).


Concurrently with our work several papers have presented results on manipulation tasks. \citet{rajeswaran2017learning,nair2017overcoming} both use human demonstrations to aid exploration. \citet{nair2017overcoming} extends the DDPGfD algorithm \citep{vecerik2017leveraging} to learn a block stacking task on a position-controlled arm in simulation. \citet{rajeswaran2017learning} use the demonstrations with a form of behavioral cloning and data augmentation to learn several complex manipulation tasks. In both cases, controllers observe a low-level state feature and these methods inherently require aligned state and action spaces with the demonstrations. In contrast, our method learns end-to-end visuomotor policies without reliance on demonstrator actions. Thus, it can utilize demonstrations when raw demonstrator actions are unknown or generated by a different body.
\citet{pinto2017asymmetric} and \citet{peng2017simtoreal} address the transfer from simulation to reality, focusing on randomizing visual appearance and robot dynamics respectively. \citeauthor{peng2017simtoreal} transfer a block-pushing policy operating from state features to a 7-DoF position controlled Fetch robotics arm. \citeauthor{pinto2017asymmetric} consider different tasks using visual input with end-effector position control. These concurrent works have each introduced a subset of techniques that our model employs. This work, developed independently from concurrent works, integrates several new techniques into one coherent method. Our experimental results demonstrate that good performances come from the synergy of these combined techniques.

\section{Model}
Our goal is to learn a visuomotor policy with deep neural networks for robot manipulation tasks. The policy takes both an RGB camera observation and a proprioceptive feature vector that describes the joint positions and angular velocities. These two sensory modalities are also available on the real robot, allowing us to train in simulation and subsequently transfer the learned policy to the robot without modifications.
Fig.~\ref{fig:architecture} provides an overview of our model. The deep visuomotor policy encodes the pixel observation with a convolutional network (CNN) and the proprioceptive feature with a multilayer perceptron (MLP). The features from these two modules are concatenated and passed to a recurrent long short term memory (LSTM) layer before producing the joint velocities (control commands). The whole network is trained end-to-end. We start with a brief review of the basics of generative adversarial imitation learning (GAIL) and proximal policy optimization (PPO). Our model extends upon these two methods for visuomotor skills.

\subsection{Background: GAIL and PPO}
Imitation learning (IL) is the problem of learning a behavior policy by mimicking a set of demonstrations. Here we assume that human demonstrations are provided as a dataset of state-action pairs $\mathcal{D}=\{(s_i,a_i)\}_{i=1\ldots N}$. Some IL methods cast the problem as one of supervised learning, i.e., behavior cloning. These methods use maximum likelihood to train a parameterized policy $\pi_\theta:\mathcal{S}\rightarrow\mathcal{A}$, where $\mathcal{S}$ is the state space and $\mathcal{A}$ is the action space, such that $\theta^{*}=\arg\max_\theta \sum_{N} \log\pi_\theta(a_i|s_i)$. The behavior cloning approach works effectively when demonstrations are abundant~\citep{ross2011reduction}. However, as robot demonstrations can be costly and time-consuming to collect, we aim for a method that can learn from a handful of demonstrations. GAIL~\citep{ho2016generative} uses demonstration data efficiently
by allowing the agent to interact with the environment and learn from its own experiences. Similar to Generative Adversarial Networks (GANs)~\citep{goodfellow2014generative}, GAIL employs two networks, a policy network $\pi_\theta: \mathcal{S}\rightarrow\mathcal{A}$ and a discriminator network $D_\psi: \mathcal{S}\times\mathcal{A}\rightarrow [0,1]$. It uses a min-max objective function similar to that of GANs:
\begin{equation}
\min_{\theta} \max_\psi\, \mathbb{E}_{\pi_{E}}[\log D_\psi(s,a)] + \mathbb{E}_{\pi_\theta}[\log(1-D_\psi(s,a))],
\end{equation}
where $\pi_E$ denotes the expert policy that generated the demonstration trajectories. This objective encourages the policy $\pi_\theta$ to have an occupancy measure close to that of the expert policy.

In this work we train $\pi_\theta$ with policy gradient methods to maximize the discounted sum of the reward function $r_{gail}(s_t,a_t) = -\log(1-D_\psi(s_t,a_t))$, clipped at a max value of 10. In continuous domains, trust region methods greatly stabilize policy training. GAIL was originally presented in combination with TRPO~\citep{schulman2015trust} for updating the policy. Recently, PPO~\citep{schulman2017proximal} has been proposed as a simple and scalable approximation to TRPO. PPO only relies on first-order gradients and can be easily implemented with recurrent networks in a distributed setting~\citep{heess2017emergence}. 
PPO implements an approximate trust region that limits the change in the policy per iteration. This is achieved via a regularization term based on the Kullback-Leibler (KL) divergence, the strength of which is adjusted dynamically depending on actual change in the policy in past iterations.

\subsection{Reinforcement and Imitation Learning Model}
\subsubsection{Hybrid IL/RL Reward}
Shaping rewards are a popular means of facilitating exploration.
Although reward shaping 
can 
be very effective it 
can also lead to suboptimal 
solutions~\citep{ng1999policy}. Hence, we design the task rewards as sparse piecewise constant functions based on the different stages of the respective tasks. For example, we define three stages for the \emph{block stacking} task, including \emph{reaching}, \emph{lifting}, and \emph{stacking}. Reward change only occurs when the task transits from one stage to another. In practice, we find  defining such a sparse multi-stage reward easier than handcrafting a dense shaping reward and less prone to producing suboptimal behaviors.
Training agents in continuous domains with sparse or piecewise constant rewards is challenging. Inspired by reward augmentation as described in~\citet{li2017inferring} and \citet{merel2017learning}, we provide additional guidance via a hybrid reward function that combines the imitation reward $r_{gail}$ with the task reward $r_{task}$:
\begin{equation}
r(s_t,a_t) = \lambda r_{gail}(s_t,a_t) + (1-\lambda) r_{task}(s_t,a_t)\,\,\lambda\in [0,1].
\label{eq:hybrid_reward}
\end{equation}
Maximizing this hybrid reward can be interpreted as simultaneous reinforcement and imitation learning, where the imitation reward encourages the policy to generate trajectories closer to demonstration trajectories, and the task reward encourages the policy to achieve high returns on the task. Setting $\lambda$ to either 0 or 1 reduces this method to the standard RL or GAIL setups. In our experiments, with a balanced contribution of these two rewards the agents can solve tasks that neither GAIL nor RL can solve alone. Further, the final agents achieve higher returns than the human demonstrations owing to the exposure to task rewards. 

\subsubsection{Leveraging Physical States in Simulation}
\label{sec:leveraging_states}

The physics simulator we employ for training exposes the full state of the system.
Even though such privileged information is unavailable on a real system, we can take advantage of it when training the policy in simulation. We propose four techniques for leveraging the physical states in simulation to stabilize and accelerate learning (1) the use of a curriculum derived from demonstration states, (2) the use of privileged information for the value function (baseline), (3) the use of object-centric features in the discriminator, and (4) auxiliary tasks. We elaborate these four techniques as follows:

\textbf{1. Demonstration as a curriculum.} The problem of exploration in continuous domains is exacerbated by the long duration of realistic tasks. Previous work indicates that shaping the distribution of start states towards states where the optimal policy tends to visit can greatly improve policy learning~\citep{kakade2002approximately,popov2017data}. We alter the start state distribution with demonstration states. We build a curriculum that contains clusters of states in different stages of a task. For instance, we define three clusters for the pouring task, including \emph{reaching the mug}, \emph{grasping the mug}, and \emph{pouring}. 
During training, with probability $\epsilon$, we then start an episode from a random initial state, and with probability $1-\epsilon$ we uniformly select a cluster and initialize the episode with a demonstration state from that cluster. 
This is possible since our simulated system is fully characterized by the physical states.

\textbf{2. Learning value functions from states.} PPO uses a learnable value function $V_\phi$ to estimate the advantage required to compute the policy gradient. During training, each PPO worker executes the policy for $K$ steps and uses the discounted sum of rewards and the value as an advantage function estimator $\hat{A}_t=\sum_{i=1}^K\gamma^{i-1}r_{t+i} + \gamma^{K-1}V_\phi(s_{t+K})-V_\phi(s_t)$, where $\gamma$ is the discount factor. As the policy gradient relies on the value function to reduce variance, it is beneficial to accelerate learning of the value function. Rather than using pixels as inputs similar to the policy network, we take advantage of the low-level physical states (e.g., the position and velocity of the 3D objects and the robot arm) to train the value $V_\phi$ with a smaller multilayer perceptron. We find that training the policy and value in two different modalities stabilizes training and reduces oscillation of the agent's performance. This technique has also been 
been proposed concurrently by \citet{pinto2017asymmetric}.

\textbf{3. Object-centric discriminator.} As for the value function, we exploit the availability of the physical states for the GAIL discriminator and provide task specific features as input.
%
We find that object-centric representations (e.g., absolute and relative positions of the objects) provide the salient and relevant signals to the discriminator. The states of the robot arm in contrast
lead the discriminator to focus on irrelevant aspects of the behavior of the controller and are detrimental for
training of the policy. Inspired by information hiding strategies used in locomotion domains~\citep{heess2016learning,merel2017learning}, our discriminator only takes the object-centric features as input while masking out arm-related information.
The construction of the object-centric representation requires a certain amount of domain knowledge of the tasks. We find that the relative positions of objects and displacements from the gripper to the objects usually provide the most informative characterization of a task. Empirically, we find that our model is not very sensitive to the particular choices of object-centric features, as long as they carry sufficient task-specific information. We provide detailed descriptions in Appendix \ref{appendix:tasks}.


\textbf{4. State prediction auxiliary tasks.} Auxiliary tasks have been shown to be effective in improving the learning efficiency and the final performance of deep RL methods~\citep{jaderberg2016reinforcement}. To facilitate learning visuomotor policies we add a state prediction layer on the top of the CNN module to predict the locations of objects from the camera observation. We use a fully-connected layer to regress the 3D coordinates of objects in the task, 
minimizing the $\ell_2$ loss between the predicted and ground-truth object locations. The auxiliary tasks are not required for our model to learn good visuomotor policies; however, adding the additional supervision can often accelerate the training of the CNN module.

\begin{figure*}[t!]
    \centering
    \includegraphics[width=1.0\linewidth]{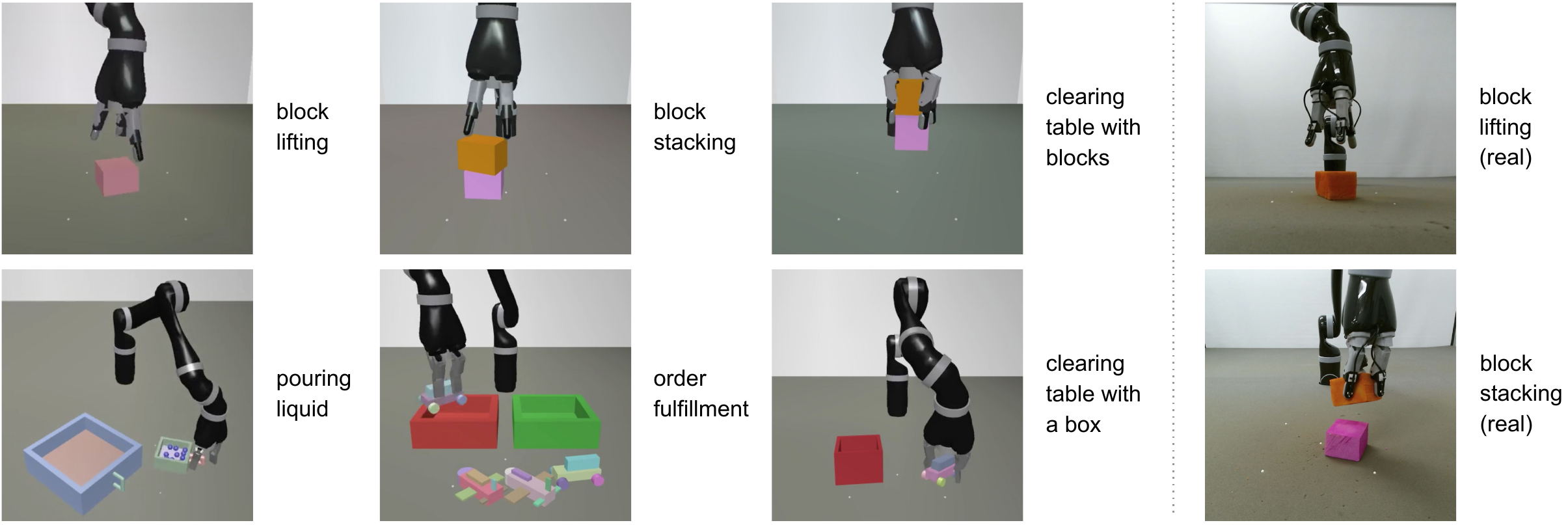}
    \caption{Visualizations of the six manipulation tasks in our experiments. The left column shows RGB images of all six tasks in the simulated environments. These images correspond to the actual pixel observations as input to the visuomotor policies. The right column shows the two tasks with color blocks on the real robot.}
    \vspace{-2mm}
    \label{fig:tasks}
\end{figure*}

\subsubsection{Sim2Real Policy Transfer}
We perform policy transfer experiments on a real-world Kinova Jaco robot arm. The simulation was manually 
adjusted
to roughly
match the
appearance 
and dynamics of the laboratory setup: a Kinect camera was visually calibrated to match the position and orientation of the simulated camera, and the simulation's dynamics parameters were manually adjusted to match the dynamics of the real arm. Instead of using professional calibration equipment, our approach to sim2real policy transfer relies on domain randomization of camera position and orientation~\citep{james2017transferring,tobin2017domain}. 
 In contrast to 
 some
 previous works our trained policies do not rely on any object position information or intermediate goals but rather learn a mapping end-to-end from raw pixel input joint velocities.
In addition, to improve the robustness of our controllers to latency effects on the real robot, we also fine-tune our policies while subjecting them to action dropping.
A detailed description is available in Appendix \ref{appendix:sim2real}. 

\section{Experiments}
Here we demonstrate that our approach offers a 
flexible
framework to visuomotor policy learning. 
To this end
we evaluate 
its performance on the
six manipulation tasks illustrated in Fig.~\ref{fig:tasks}. We provide additional qualitative results in this {\color{RoyalBlue} \href{https://youtu.be/EDl8SQUNjj0}{video}}.

\begin{figure*}[t!]
    \centering
    \begin{subfigure}[t]{0.325\linewidth}
        \centering
        \includegraphics[width=1.0\linewidth]{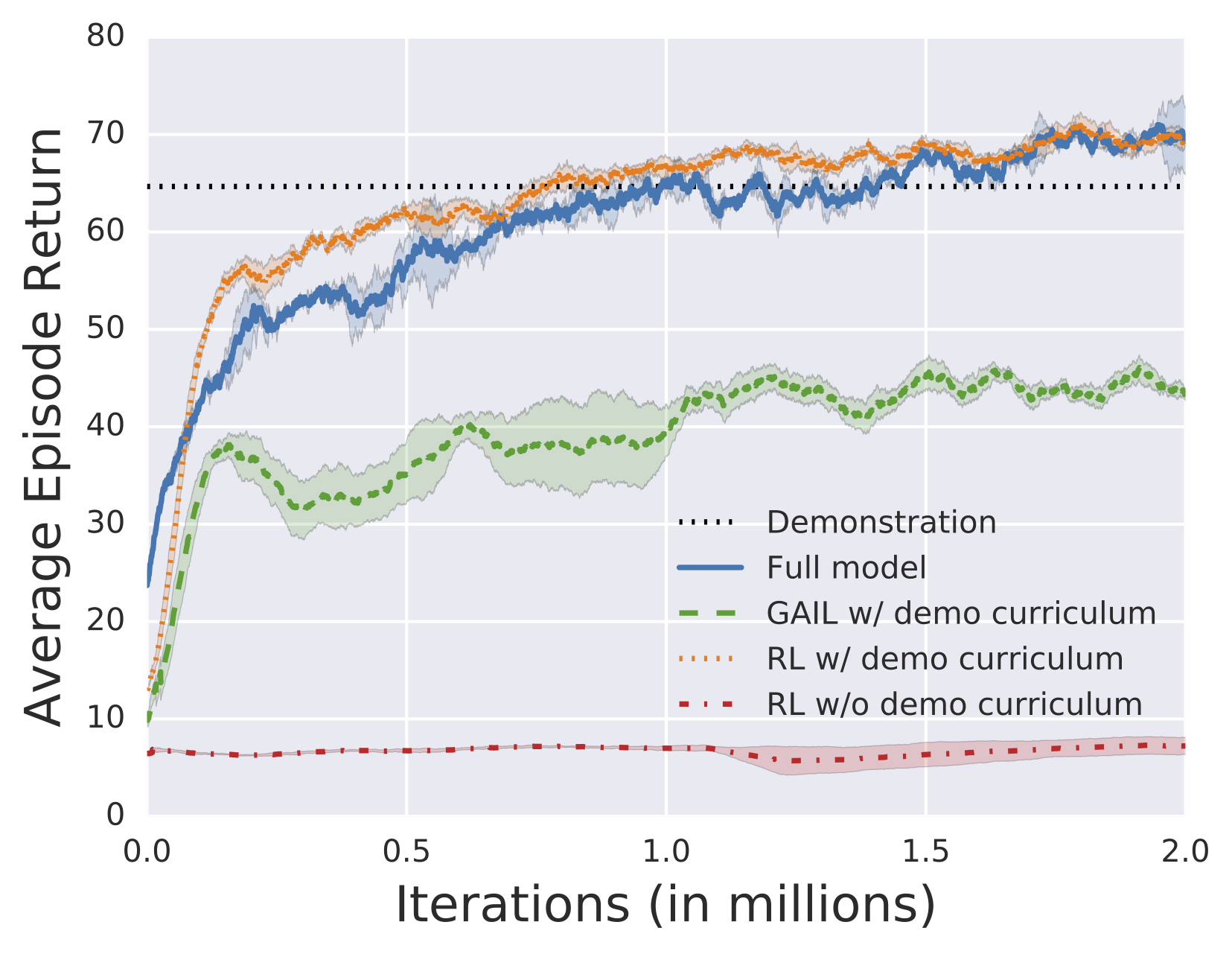}
        \caption{Block lifting}
    \end{subfigure}
    \begin{subfigure}[t]{0.325\linewidth}
        \centering
        \includegraphics[width=1.0\linewidth]{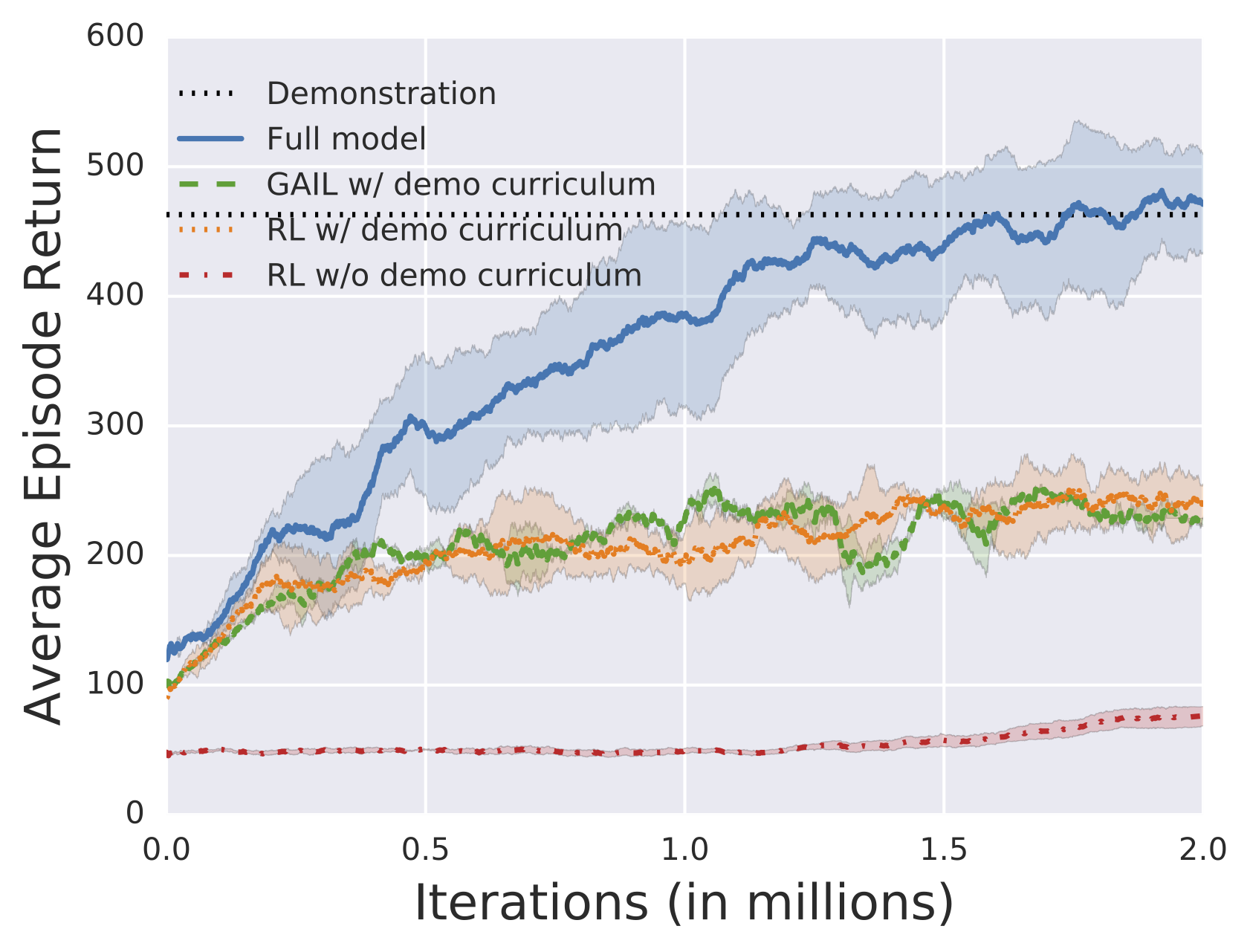}
        \caption{Block stacking}
    \end{subfigure}
    \begin{subfigure}[t]{0.325\linewidth}
        \centering
        \includegraphics[width=1.0\linewidth]{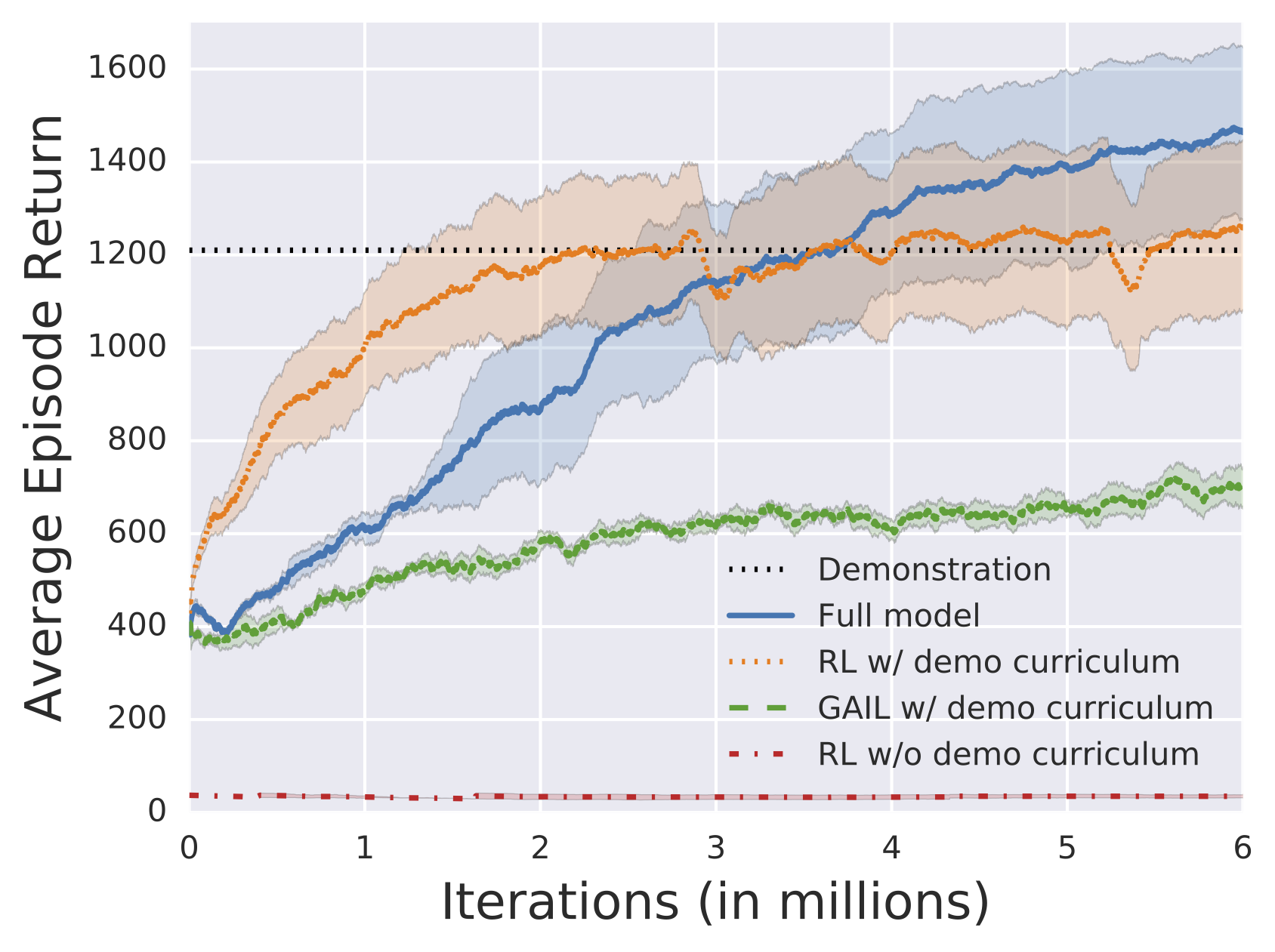}
        \caption{Clearing table with blocks}
    \end{subfigure}
    \begin{subfigure}[t]{0.325\linewidth}
        \centering
        \includegraphics[width=1.0\linewidth]{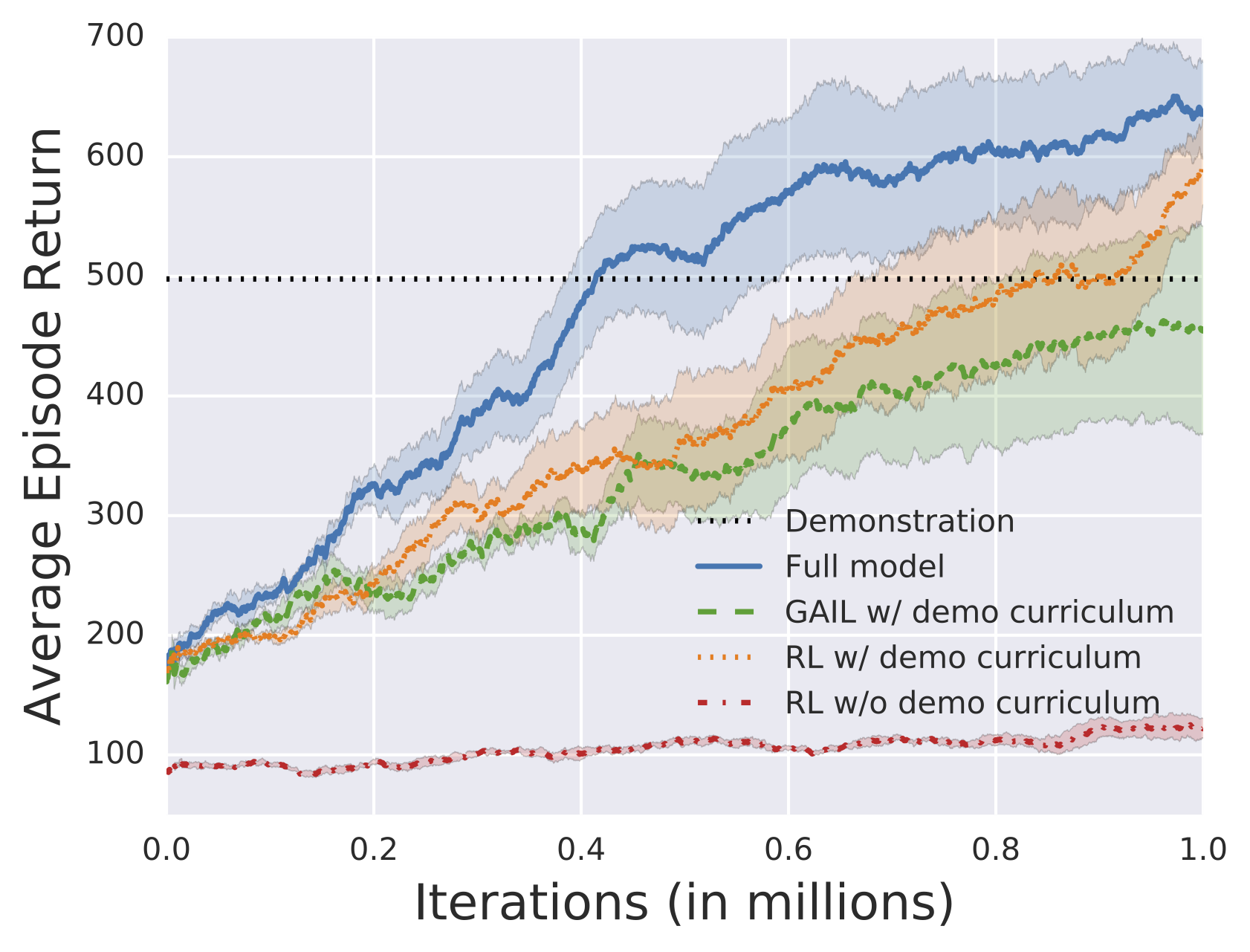}
        \caption{Clearing table with a box}
    \end{subfigure}
    \begin{subfigure}[t]{0.325\linewidth}
        \centering
        \includegraphics[width=1.0\linewidth]{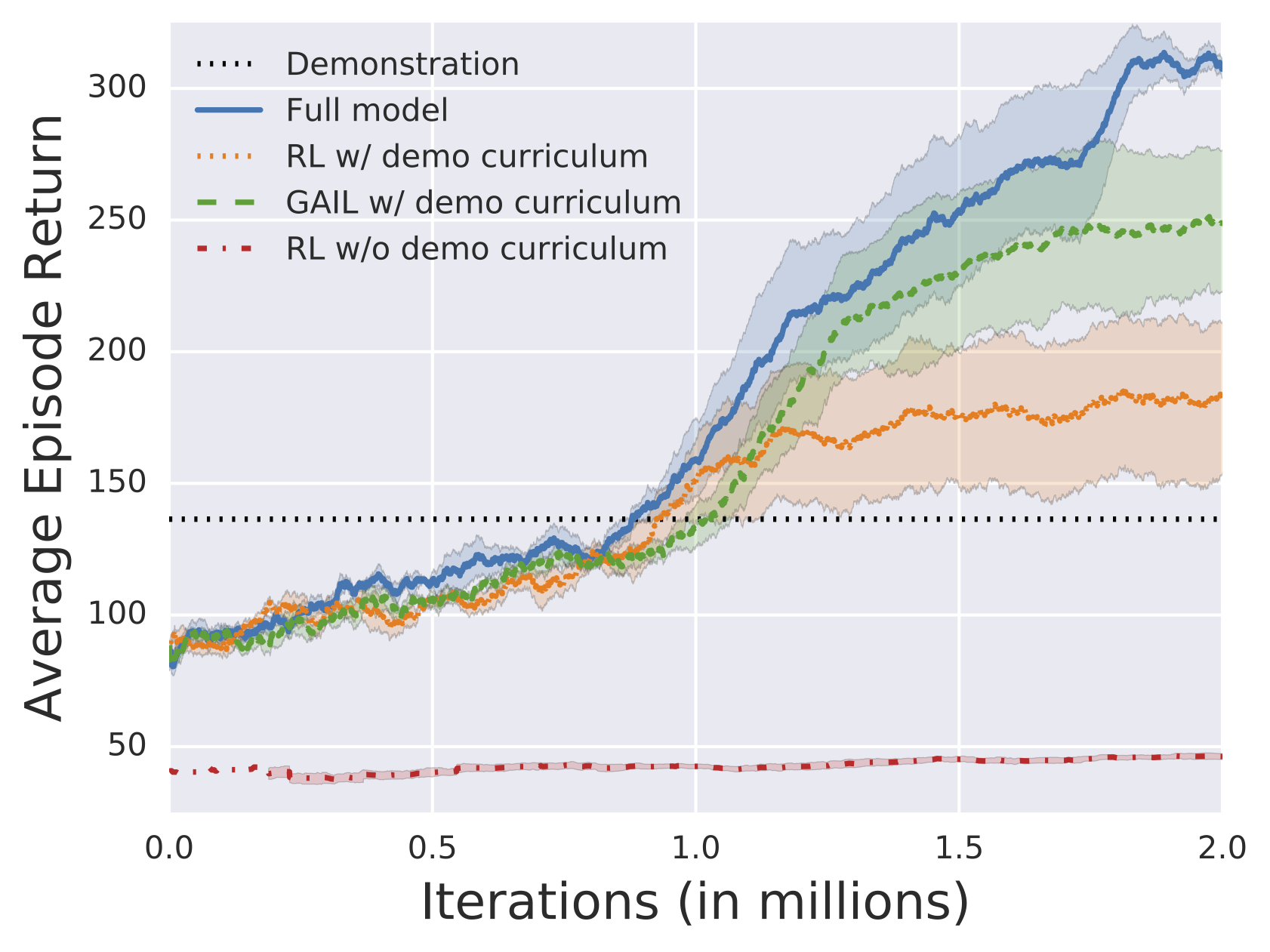}
        \caption{Pouring liquid}
    \end{subfigure}
    \begin{subfigure}[t]{0.325\linewidth}
        \centering
        \includegraphics[width=1.0\linewidth]{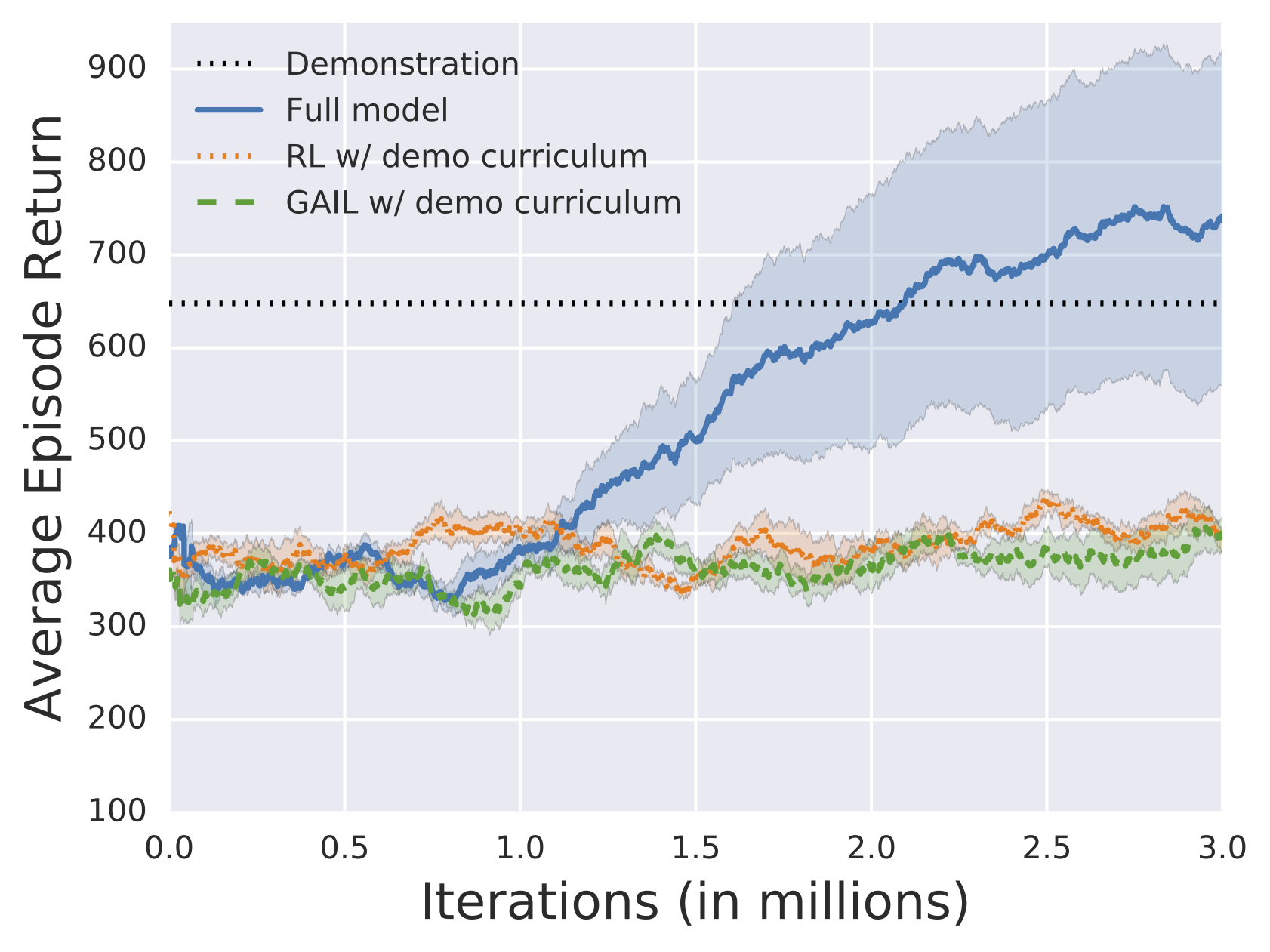}
        \caption{Order fulfillment}
    \end{subfigure}
    \caption{Learning efficiency of our reinforcement and imitation model against baselines. The plots are averaged over 5 runs with different random seeds. All the policies use the same network architecture and the same hyperparameters (except $\lambda$).}
    \label{fig:quantitative}
    \vspace{-2mm}
\end{figure*}

\subsection{Environment Setup}

We use a Kinova Jaco arm that has 9 degrees of freedom: 
six arm joints and three actuated fingers. The robot arm interacts with a diverse set of objects on a tabletop. The visuomotor policy controls the robot 
by setting the
joint velocity commands,
producing 9-dimensional continuous velocities in the range of $[-1, 1]$ at 20Hz. The proprioceptive features consist of the positions and angular velocities of the arm joints and the fingers. 
Visual observations of the table-top scene are provided via a suitably positioned real-time RGB camera.
The proprioceptive features and the camera observations are available in both simulation and real environments thus enabling policy transfer.
%
The physical environment is simulated in the MuJoCo physics simulator~\citep{todorov2012mujoco}.

We use a large variety of objects, ranging from basic geometric shapes to procedurally generated 3D objects built from ensembles of primitive shapes. We increase the diversity of objects by randomizing various physical properties, including dimension, color, mass, friction, etc. We collect demonstrations using a SpaceNavigator 3D motion controller, which allows us to operate the robot arm with a position controller, and gather 30 episodes of demonstration for each task including observations, actions, and physical states. As each episode takes less than a minute to complete, demonstrating each task can be done within half an hour. 

\subsection{Robot Arm Manipulation Tasks}

Fig.~\ref{fig:tasks} shows the six manipulation tasks in our experiments. The first column shows the six tasks in simulated environments, and the second column shows the real-world setup of the block lifting and stacking tasks. We see obvious visual discrepancies of the same task in simulation and reality. These six tasks exhibit learning challenges to varying degrees. 
The first three tasks use simple colored blocks, which 
makes it easy to replicate a similar setup on the real robot.
We study sim2real policy transfer with the block lifting and stacking tasks in Sec.~\ref{sec:sim2real_results}.

\textbf{Block lifting.} The goal is to grasp and lift a randomized block, allowing us to evaluate the model's robustness. We vary several random factors, including the robot arm dynamics  (friction and armature), lighting conditions, camera poses, background colors, as well as the properties of the block. Each episode starts with a new configuration with these random factors uniformly drawn from a preset range.

\textbf{Block stacking.} The goal is to stack one block on top of the other block. Together with the block lifting task, this is evaluated in sim2real transfer experiments.

\textbf{Clearing table with blocks.} This task requires lifting two blocks off the tabletop. 
One solution is to stack the blocks and lift them both together. This task requires longer time and a more dexterous controller, introducing a significant challenge for exploration. 

The next three tasks involve a large variety of procedurally generated 3D shapes, making them difficult to recreate in real environments. We use them to examine the model's ability to generalize across object variations in long and complex tasks.

\textbf{Clearing table with a box.} The goal is to clear the tabletop that has a box and a toy car. One strategy is to grasp the toy, put it into the box, and lift the box. Both the box and the toy car are randomly generated for each episode.

\textbf{Pouring liquid.} Modeling and reasoning about deformable objects and fluids is a long-standing challenge in the robotics community~\citep{schenck2017reasoning}. We design a pouring task where we use many small spheres to simulate liquid. The goal is to pour the ``liquid'' from one mug to the other container. This task is particularly challenging due to the dexterity required. Even humans struggled to demonstrate the task with our 3D motion controller after extensive practice.

\textbf{Order fulfillment.} In this task we randomly place a variable number of procedurally generated toy planes and cars on the table. The goal is to place all the planes into the green box and all the cars into the red box. This task requires the policy to generalize at an abstract level. It needs to recognize the object categories, perform successful grasps on diverse shapes, and handle tasks with variable lengths.

\begin{figure*}[t!]
    \centering
    \begin{subfigure}[t]{0.49\linewidth}
        \centering
        \includegraphics[width=1.0\linewidth]{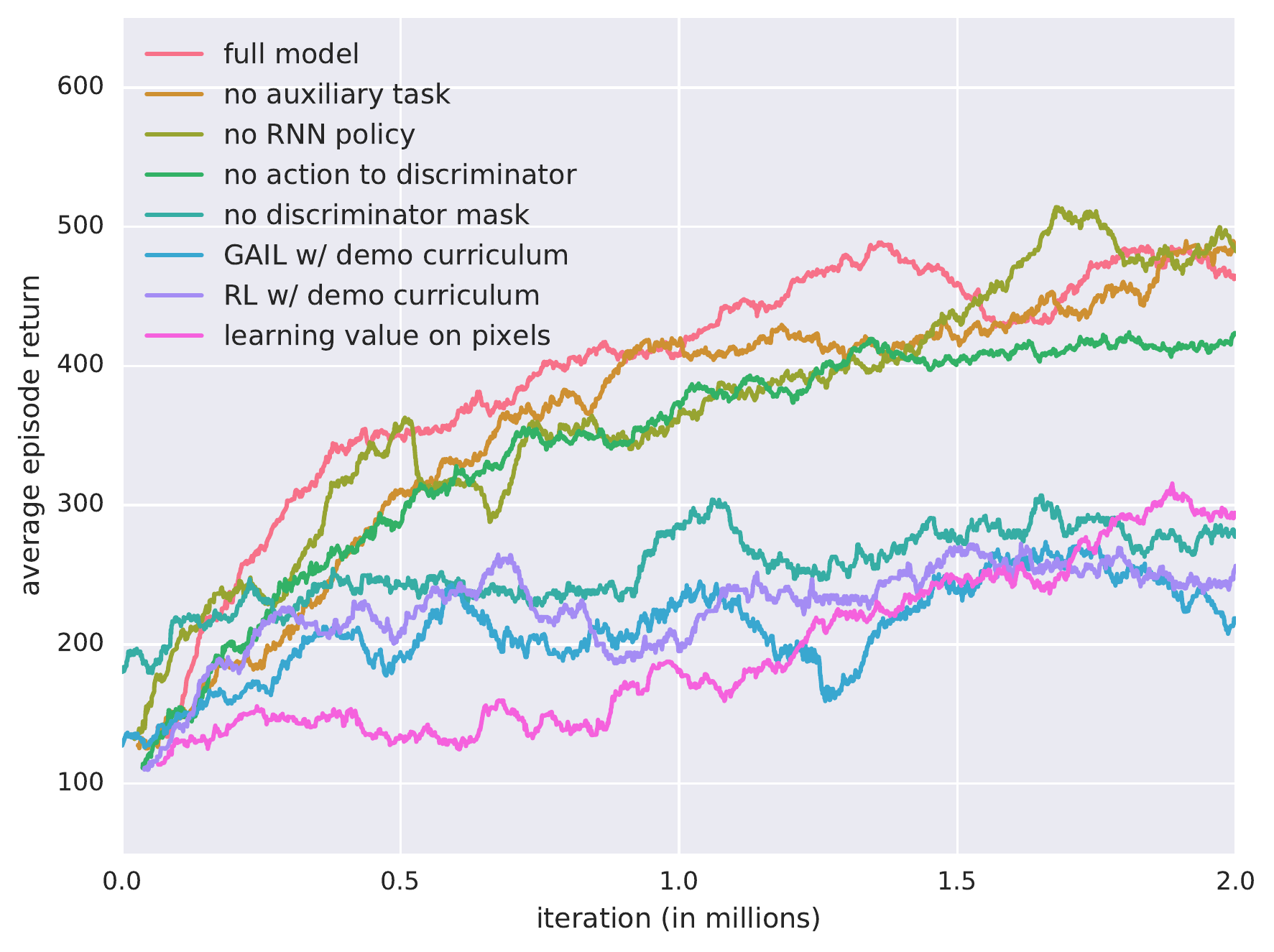}
        \caption{Ablation study of model components}
        \label{fig:stack_ablation}
    \end{subfigure}
    \begin{subfigure}[t]{0.49\linewidth}
        \centering
        \includegraphics[width=1.0\linewidth]{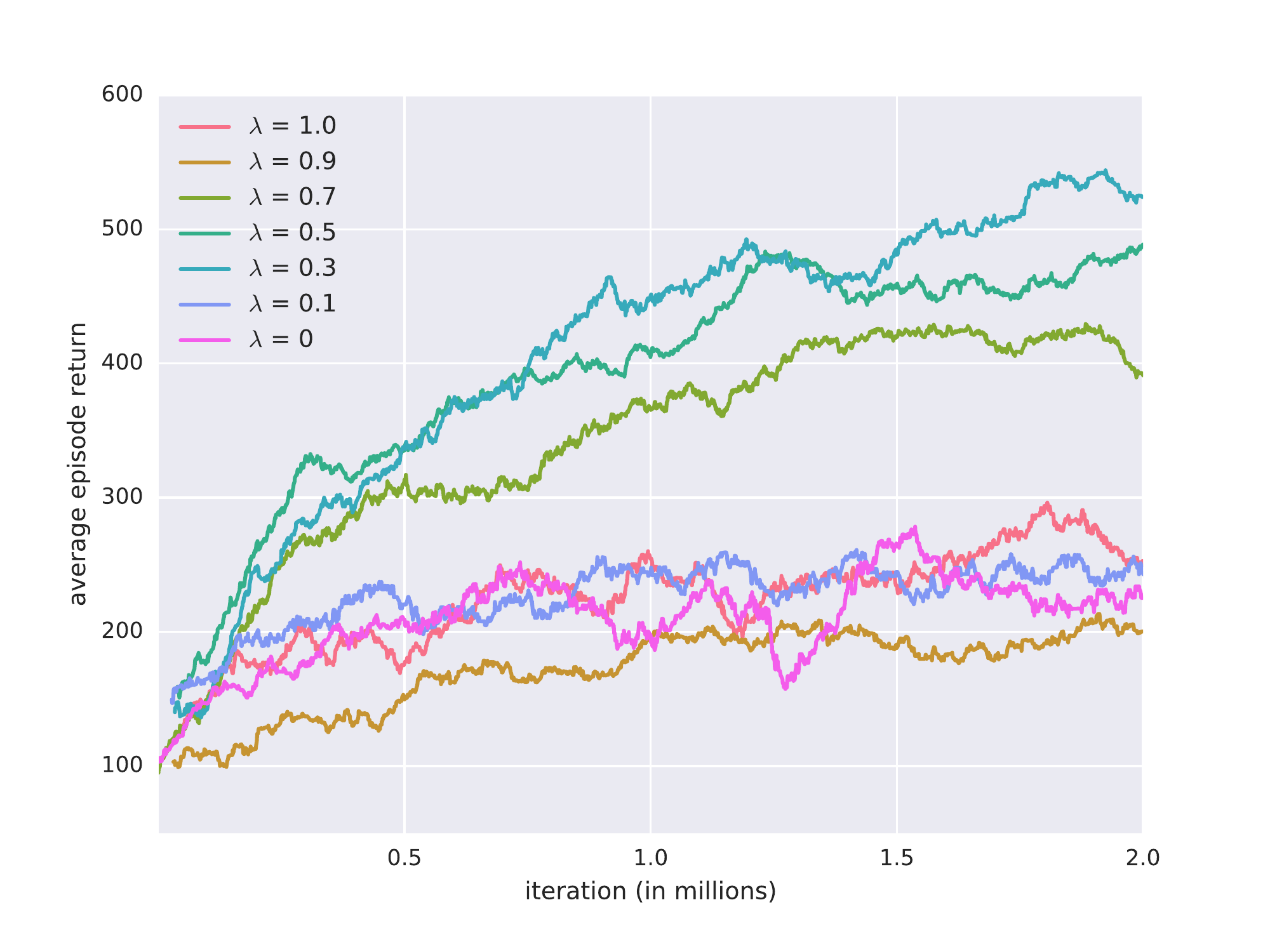}
        \caption{Model sensitivity to $\lambda$ values}
        \label{fig:stack_lambda}
    \end{subfigure}
    \caption{Model analysis in the stacking task. On the left we investigate the impact on performance by removing each individual component from the full model. On the right we investigate the model's sensitivity to the hyperparameter $\lambda$ that moderates the contribution of reinforcement and imitation.}
    \label{fig:stack_task_analysis}
    \vspace{-2mm}
\end{figure*}

\subsection{Quantitative Evaluation}

Our full model can solve all six tasks, with only occasional failures, using the same policy network, the same training algorithm, and a fixed set of hyperparameters. On the contrary, neither reinforcement nor imitation alone can solve all tasks. We compare the full model with three baselines
which correspond to pure RL, pure GAIL, and RL w/o demonstration curriculum. These baselines use the same setup as the full model, except that we set $\lambda=0$ for RL and $\lambda=1$ for GAIL, while our model uses a balanced contribution of the hybrid reward, where $\lambda=0.5$. In the third baseline, all training episodes start from random initial states rather than resetting to demonstration states. This is a standard RL setup.

We report the mean episode returns as a function of the number of training iterations in Fig.~\ref{fig:quantitative}. Our full model achieves the highest returns in all six tasks. The only case where the baseline model is on par with the full model is the block lifting task, in which both the RL baseline and the full model achieved similar levels of performance. We hypothesize that this is due to the short length of the lifting task, where random exploration 
can provide a sufficient learning signal without the aid of demonstrations. 
In the other five tasks, the full model outperforms both the reinforcement learning and imitation learning baselines by a large margin, demonstrating the effectiveness of combining reinforcement and imitation for learning complex tasks. Comparing the two variants of RL with and without using demonstration as a curriculum, we see a pronounced effect of altering the start state distribution. We see that RL from scratch leads to very slow learning progress; while initiating episodes along demonstration trajectories enables the agent to train on states from different stages of a task. As a result, it greatly reduces the burden of exploration and improves the learning efficiency. We also report the mean episode returns of human demonstrations in these figures. 
Demonstrations with the
3D motion controller are imperfect, especially for pouring (see {\color{RoyalBlue} \href{https://www.youtube.com/watch?v=EDl8SQUNjj0&feature=youtu.be&t=2m43s}{video}}), 
and the trained agents exceed the performance of the human operator.

Two findings are noteworthy. First, the RL agent learns faster than the full model in the clearing blocks task, but the full model eventually outperforms. This is because the full model discovers a novel strategy, different from the strategy employed by human operators (see {\color{RoyalBlue} \href{https://www.youtube.com/watch?v=EDl8SQUNjj0&feature=youtu.be&t=3m0s}{video}}).
In this case, imitation gave contradictory signals but eventually, reinforcement learning guided the policy towards a better strategy. Second, pouring liquid is the only task where GAIL outperforms its RL counterpart. Imitation can effectively shape the agent's behaviors towards the demonstration trajectories~\citep{wang2017robust}. This is a viable solution for the pouring task, where a controller that generates similar-looking behaviors can complete the task. In contact-rich domains with sufficient variation, however, a controller 
trained only from a small number of demonstrations will struggle to handle the complex dynamics and to generalize appropriately to novel instances of the task.
We hypothesize that this is why the baseline RL agent outperforms the GAIL agent in the other five tasks.

We further perform an ablation study on the block stacking task to understand the impact of different components of our model. In Fig.~\ref{fig:stack_ablation}, we trained our agents with a number of configurations, each with a single modification to the full model. We see that these
ablations cluster
into two groups: agents that learn to stack (with average returns greater than 400) and agents that only learn to lift (with average returns between 200 and 300). These results indicate that the hybrid RL/IL reward, learning value function from states, and object-centred features for the discriminator play an integral role in learning good policies. Using only the RL or GAIL reward, learning the value function from pixels, or providing the full arm state as discriminator input (no discriminator mask) all result in inferior performance. In contrast, the optional components include the recurrent policy core (LSTM), the use of state prediction auxiliary tasks, and whether to include actions in discriminator input. 
This result suggests that our model can learn end-to-end visuomotor policies without a pretraining phase or the need of auxiliary tasks, as opposed to previous work on visuomotor learning \citep{chebotar2017path,levine2015end,yahya2016collective}.
Furthermore, it can work when the GAIL discriminator only has access to the demonstration states without the accompanying demonstrator actions. 
Therefore, it can potentially use demonstrations 
collected with a different body
where the underlying controls are unknown or different from the robot's actuators.
We then examine the model's sensitivity to the $\lambda$ values in Eq.~\ref{eq:hybrid_reward}. We see in Fig.~\ref{fig:stack_lambda} that, our model works well with a broad range of $\lambda$ values from 0.3 to 0.7 that provide a balanced mix of the RL and GAIL rewards. 

\subsection{Sim2Real Policy Transfer Results}
\label{sec:sim2real_results}
To assess the robustness of the simulation-trained policy, we evaluate zero-shot transfer (no additional training) on a real Jaco arm. 
The real-world setup was roughly matched to the simulation environment
(including camera positions and robot kinematics and approximate object size and color).
We execute the trained policy network on the robot and count the number of successful trials for both the lifting and stacking tasks. The arm position 
is 
randomly initialized and the target block(s) are placed in a number of repeatable start configurations for each task. The zero-shot transfer of the lifting policy 
has
a success rate of $64\%$ over $25$ trials (split between $5$ block configurations). The stacking policy 
has a success rate of $35\%$ over 20 trials (split between $2$ block configurations). $80\%$ of the stacking trajectories, however, contain successful lifting behavior, and $100\%$  contains successful reaching behavior. It is impractical to conduct a fair comparison with previous work~\cite{james2017transferring,tobin2017domain,viereck2017learning} that implemented different tasks and different configurations. The state-of-the-art sim2real work closest to our setup is progressive network~\citep{rusu2016progressive,rusu2016simtoreal}, which has demonstrated block reaching behaviors with a pixel-to-action RL policy on a Jaco arm. Their work did not demonstrate any lifting or stacking behavior, while our method has achieved reaching behaviors with a $100\%$ success rate. Qualitatively, the policies are notably robust even on failed attempts. The stacking policy repeatedly chases the block to get a successful grasp before trying to stack (see {\color{RoyalBlue} \href{https://youtu.be/EDl8SQUNjj0?t=3m23s}{video}}). For more detailed descriptions of the sim2real results, refer to Appendix \ref{appendix:sim2real}. 

Several aspects of system mismatch have constrained the policies from attaining a better performance  on the real robot. Although the sim and real domains 
are
similar, there 
is
still a sizable reality gap that 
makes
zero-shot transfer challenging. For example, while the simulated blocks are rigid the objects employed in the real-world setup are
non-rigid foam blocks which 
deform and bounce unpredictably. 
Furthermore, neural network policies are sensitive to subtle discrepancies between simulated rendering and the real camera frame. Nonetheless, the preliminary successes achieved by these policies offer a good starting point for future work to leverage a small amount of real-world experience to enable better transfer.

\section{Discussion}

In this paper, we have described a 
general model-free deep reinforcement learning method for end-to-end learning of policies that operate from RGB camera images and perform manipulation using joint velocity control.
Our method 
combines the use of demonstrations via generative adversarial imitation learning~\cite{ho2016generative} with model-free RL to achieve both effective learning of difficult tasks and robust generalization.
%
The approach only requires a small number of demonstration trajectories  (30 per task in the experiments). Additionally, this approach works from state trajectories (without demonstrator actions) combined with the use of only partial/featurized demonstrations being seen by the discriminator -- this can simplify and increase the flexibility during data collection and facilitate generalization beyond conditions seen in the demonstrations (e.g. demonstrations could potentially be collected with a different body, such as a human demonstrator via motion capture). Demonstrations were collected via teleoperation of the simulated arm in less than thirty minutes per task. Our method integrates several new techniques to leverage the flexibility and scalability afforded by simulation, such as access to privileged information and the use of large-scale RL algorithms. The experimental results have demonstrated its effectiveness in complex manipulation tasks in simulation and achieved preliminary successes of zero-shot transfer to real hardware. We trained all the policies with the same policy network, the same training algorithm, and the same hyperparameters. 
The approach makes some use of task-specific information especially in the choice of the object-centric features for the discriminator and the RL reward. In practice we have found the specification of these features intuitive, and our method was reasonably robust to specific choices, thus striking favorable balance between the need for (limited) prior knowledge and the 
generality of the solutions that can be learned for complex tasks.


In order to fulfill the potential of deep RL in robotics, it is essential to confront the full variability of the real-world, including diversity of object appearances, system dynamics, task semantics, etc. We have therefore focused on learning controllers that could handle significant task variations along multiple dimensions. 
To improve a policy's ability to generalize, we have increased the diversity of training conditions with parameterized, procedurally generated 3D objects and randomized system dynamics. This has resulted in policies that exhibit robustness to large variations in simulation as well as against some of the domain discrepancy between simulation and the real world.

Simulation is at the center of our method. Training in simulation circumvents several practical challenges of deep RL for robotics, such as access to state information for reward specification, high sample complexity, and safety considerations. Training in simulation also allows us to use the simulation state to facilitate and stabilize training (i.e. by providing state information to the value function), which in our experiments has been important for learning good visuomotor policies. However, even though our method utilizes such privileged information during training 
it ultimately produces policies that only rely on vision and proprioceptive information of the arm and 
that can thus be deployed on real hardware. 

Executing the policies on the real robot reveals  
that there remains a sizable domain gap between simulation and real hardware. Transfer is affected by visual discrepancies as well as by differences in the arm dynamics and in the physical properties of the environment. 
This 
leads to a certain level of performance degradation when running a simulation policy on the real robot. Still, our real-world experiments have exemplified that zero-shot sim2real transfer can achieve initial success with RL trained policies performing 
pixel-to-joint-velocity control.
%

\section{Conclusion}
\label{sec:Discussion}

We have shown that combining reinforcement and imitation learning considerably improves our ability to train systems capable of solving challenging dexterous manipulation tasks from pixels. Our method implements all three stages of a 
pipeline for robot skill learning: first, we collected a small amount of demonstration data to simplify the exploration problem; second, we relied on physical simulation to perform large-scale distributed robot training; and third, we performed sim2real transfer for real-world deployment. In future work, we seek to improve the sample efficiency of the learning method and to leverage real-world experience to close the reality gap for policy transfer.
\section*{Acknowledgment}
The authors would like to thank Yuval Tassa, Jonathan Scholz, Thomas Roth\"{o}rl, Jonathan Hunt, and many other colleagues at DeepMind for the helpful discussion and feedback.

\bibliographystyle{plainnat}
\bibliography{iclr2018_conference}

\clearpage
\newpage
\appendices
\section{Experiment Details}

The policy network takes the pixel observation and the proprioceptive feature as input. The pixel observation is an RGB image of size $64\times 64\times 3$. We used the Kinect for Xbox One camera\footnote{\url{https://www.xbox.com/en-US/xbox-one/accessories/kinect}} in the real environment. The proprioceptive feature describes the joint positions and velocities of the Kinova Jaco arm.\footnote{\url{http://www.kinovarobotics.com}} Each joint position is represented as the $\sin$ and $\cos$ of the angle of the joint in joint coordinates. Each joint velocity is represented as the scalar angular velocity. This results in a 24-dimensional proprioceptive feature that contains the positions (12-d) and velocities (6-d) of the six arm joints and the positions (6-d) of the three fingers. We exclude the finger velocities due to the noisy sensory readings on the real robot. When collecting demonstrations, we use a 6-DoF SpaceNavigator motion controller\footnote{\url{https://www.3dconnexion.com/spacemouse_compact}} to command the end effector to complete the tasks.

We used Adam~\citep{kingma2014adam} to train the neural network parameters. We set the learning rate of policy and value to $10^{-4}$ and $10^{-3}$ respectively, and $10^{-4}$ for both the discriminator and the auxiliary tasks. The pixel observation is encoded by a two-layer convolutional network.
We use 2 convolutional layers followed by
a fully-connected layer with $128$ hidden units. The first convolutional layer has
$16$ $8 \times 8$ filters with stride 4 and the second 32 $4 \times 4$ filters with
stride $2$. We add a recurrent layer of 100 LSTM units before the policy and value outputs. The policy output is the mean and the standard deviation of a conditional Gaussian distribution over the 9-dimensional joint velocities. The initial policy standard deviation is set to $\exp(-3)$ for the \textit{clearing table with blocks} task and $\exp(-1)$ for the other five tasks. The auxiliary head  of the policy contains a separate three-layer MLP sitting on top of the convolutional network.
The first two layers of the MLP has $200$ and $100$ hidden units respectively,
while the third layer predicts the auxiliary outputs.
Finally, the discriminator is a simple three-layer MLP of $100$ and $64$ hidden units
for the first two layers with the third layer producing log probabilities.
The networks use $\tanh$ nonlinearities.

We trained the visuomotor policies using the distributed PPO algorithm~\citep{heess2017emergence} with synchronous gradient updates from 256 CPU workers. Each worker runs the policy to complete an entire episode before the parameter updates are computed. We set a constant episode length for each task based on its difficulty, with the longest being 1000 time steps (50 seconds) for the \textit{clearing table with blocks} and \textit{order fulfillment} tasks. We set $K=50$ as the number of time steps for computing $K$-step returns and truncated backpropagation through time to train the LSTM units. After a worker collects a batch of data points, it performs $50$ parameter updates for the policy and value networks, $5$ for the discriminator and $5$ for the auxiliary prediction network.


\section{Sim2Real Details}
\label{appendix:sim2real}
\begin{figure*}[t!]
    \centering
    \includegraphics[width=.8\linewidth]{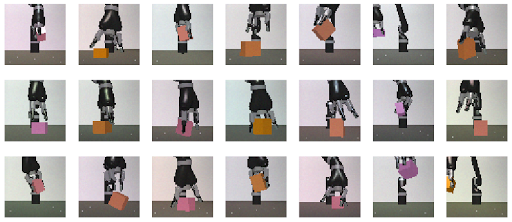}
    \caption{Tiles show the representative range of diversity seen in the domain-randomized variations of the colors, lighting, background, etc.}
    \label{fig:domain_randomizations}
    \vspace{-2mm}
\end{figure*}

To better facilitate sim2real transfer, we lower the frequency at which we sample the observations.
Pixel observations are only observed at the rate of 5Hz despite the fact that our controller runs at 20Hz.
Similarly, the proprioceptive features are observed at a rate of 10Hz.
In addition to observation delays, we also apply domain variations.
Gaussian noise (of standard deviation 0.01) are added proprioceptive features.
Uniform integers noise in the range of $[-5, 5]$ are added to each pixel independently.
Pixels of values outside the range of $[0, 255]$ are clipped.
We also vary randomly the shade of grey on the Jaco arm, the color of the table top, 
as well as the location and orientation of the light source (see Fig. \ref{fig:domain_randomizations}).

In the case of block lifting, we vary in addition the dynamics of the arm.
Specifically, we dynamically change the friction, damping, armature, and gain parameters of the robot arm
in simulation to further enhance the agent's robustness.

\begin{table}[t]
\caption{Block lifting success rate from different positions ({\bf LL, LR, UL, UR, and C} represent the positions of lower left, lower right, upper left, upper right, and center respectively).}
\label{tb:lift_comp}
\begin{center}
\begin{tabular}{l|r|r|r|r|r|r}
 \hline
\multicolumn{1}{c}{\bf } &\multicolumn{1}{r}{\bf LL} &\multicolumn{1}{r}{\bf LR}&\multicolumn{1}{r}{\bf UL} &\multicolumn{1}{r}{\bf UR} &\multicolumn{1}{r}{\bf C} &\multicolumn{1}{r}{\bf All}
\\ \hline
No Action Dropping  &  2/5  & 2/5 &  1/5  & 3/5  & 4/5 & 12/25 \\
Action Dropping  &  4/5  & 4/5 &  4/5  & 0/5 & 4/5 & 16/25 \\
\end{tabular}
\vspace{-2mm}
\end{center}
\end{table}

\begin{table}[t]
\caption{Success rate of the block stacking agent (with action dropping) from different starting positions ({\bf Left} and {\bf Right} indicate the positions of the support block upon initialization).
}
\label{tb:stack_comp}
\begin{center}
\begin{tabular}{l|r|r|r}
 \hline
\multicolumn{1}{c}{\bf } &\multicolumn{1}{r}{\bf Left} &\multicolumn{1}{r}{\bf Right}&\multicolumn{1}{r}{\bf All}
\\ \hline
Stacking Success Rate  &  5/10  & 2/10 &  7/20  \\
Lifting Success Rate  &  9/10  & 7/10 &  16/20  \\
\end{tabular}
\vspace{-5mm}
\end{center}
\end{table}

\subsection{Action Dropping}
Our analysis indicates that, on the real robot, there is often a delay in the execution of actions.
The amount of delay also varies significantly.
This has an adverse effect on the performance of our agent on the physical robot
since our agents' performance depends on the timely execution of their actions.
To better facilitate the transfer to the real robot, we fine-tune our
trained agent in simulation while subjecting them to a random chance of dropping actions.
Specifically, each action emitted by the agent has a $50\%$ chance of being executed immediately
in which case the action is flagged as the last executed action.
If the current action is not executed, the last executed action
will then be executed.
Using the above procedure, we fine-tune our agents on both block lifting and block stacking
for a further $2$ million iterations.

To demonstrate the effectiveness of action dropping, we compare our agent on
the real robot over the task of block lifting.
Without action dropping, the baseline agent lifts $48\%$ percent of the time.
After fine-tuning using action dropping, our agent succeeded $64\%$ percent of the time.
For the complete set of results, please see Table~\ref{tb:lift_comp} and Table~\ref{tb:stack_comp}.

\section{Task Details} 
\label{appendix:tasks}

We use a fixed episode length for each task, which is determined by the amount of time a skilled human demonstrator can complete the task. An episode terminates when a maximum number of agent steps are performed. The robot arm operates at a control frequency of 20Hz, which means each time step takes 0.05 second.

We segment into a sequence of stages that represent an agent's progress in a task. For instance, the \emph{block stacking} task can be characterized by three stages, including \emph{reaching the block}, \emph{lifting the block} and \emph{stacking the block}. We define functions on the underlying physical state to determine the stage of a state. This way, we can cluster demonstration states according to their corresponding stages. These clusters are used to reset training episodes in our demonstration as a curriculum technique proposed in Sec.~\ref{sec:leveraging_states}. 
The definition of stages also gives rise to a convenient way of specifying the reward functions without hand-engineering a shaping reward. We define a piecewise constant reward function for each task, where we assign the same constant reward to all the states that belong to the same stage. We detail the stages, reward functions, auxiliary tasks, and object-centric features for the six tasks in our experiments.

\textbf{Block lifting.} Each episode lasts 100 time steps. We define three stages and their rewards (in parentheses) to be initial (0), reaching the block (0.125) and lifting the block (1.0). The auxiliary task is to predict the 3D coordinates of the color block. The object-centric feature consists of the relative position between the gripper and the block.

\textbf{Block stacking.} Each episode lasts 500 time steps. We define four stages and their rewards to be initial (0), reaching the orange block (0.125), lifting the orange block (0.25), and stacking the orange block onto the pink block (1.0). The auxiliary task is to predict the 3D coordinates of the two blocks. The object-centric feature consists of the relative positions between the gripper and the two blocks respectively.

\textbf{Clearing table with blocks.} Each episode lasts 1000 time steps. We define five stages and their rewards to be initial (0), reaching the orange block (0.125), lifting the orange block (0.25), stacking the orange block onto the pink block (1.0), and lifting both blocks off the ground (2.0). The auxiliary task is to predict the 3D coordinates of the two blocks. The object-centric feature consists of the 3D positions of the two blocks as well as the relative positions between the gripper and the two blocks respectively.

\textbf{Clearing table with a box.} Each episode lasts 500 time steps. We define five stages and their rewards to be initial (0), reaching the toy (0.125), grasping the toy (0.25), putting the toy into the box (1.0), and lifting the box (2.0). The auxiliary task is to predict the 3D coordinates of the toy and the box. The object-centric feature consists of the 3D positions of the toy and the box as well as the relative positions between the gripper and these two objects respectively.

\textbf{Pouring liquid.} Each episode lasts 500 time steps. We define three stages and their rewards to be initial (0), grasping the mug (0.05), pouring ($0.1N$), where $N$ is the number of small spheres in the other container. The auxiliary task is to predict the 3D coordinates of the mug. The object-centric feature consists of the 3D positions of the mug, the relative position between the gripper and the mug, and the relative position between the mug and the container.

\textbf{Order fulfillment.} Each episode lasts 1000 time steps. The number of objects varies from 1 to 4 across episodes. We define five stages that correspond to the number of toys in the boxes. The immediate reward corresponds to the number of toys placed in the correct boxes (number of toy planes in the green box and toy cars in the red box). To handle the variable number of objects, we only represent the objects nearest to the gripper for the auxiliary task and the object-centric feature. The auxiliary task is to predict the 3D coordinates of the nearest plane and the nearest car to the gripper. The object-centric feature consists of the relative positions from the gripper to these two nearest objects.


\end{document}